\documentclass[conference]{IEEEtran}
\IEEEoverridecommandlockouts
\usepackage{cite}

\usepackage{amsmath,amssymb,amsfonts}
\usepackage{algorithmic}
\usepackage{graphicx}
\usepackage{textcomp}
\usepackage{xcolor}
\def\BibTeX{{\rm B\kern-.05em{\sc i\kern-.025em b}\kern-.08em
    T\kern-.1667em\lower.7ex\hbox{E}\kern-.125emX}}
    
\makeatletter

\def\ps@IEEEtitlepagestyle{%
  \def\@oddfoot{\mycopyrightnotice}%
  \def\@evenfoot{}%
}
\def\mycopyrightnotice{%

  {\footnotesize 978-0-7381-1102-5/20/\$31.00~\copyright~2020 IEEE\hfill}
  \gdef\mycopyrightnotice{}
}

\begin{document}

\title{Deep Learning Approach Combining Lightweight CNN Architecture with Transfer Learning: An Automatic Approach for the  Detection and Recognition of Bangladeshi Banknotes}

\newcommand{\MYfooter}{\smash{
\hfil\parbox[t][\height][t]{\textwidth}{\centering
\thepage}\hfil\hbox{}}}

\makeatletter
\def\ps@IEEEtitlepagestyle{%
\def\@oddhead{\parbox[t][\height][t]{\textwidth}{\leftmark
2020 11th International Conference on Electrical and Computer Engineering (ICECE)\\
}\hfil\hbox{}}%
\def\@evenhead{\scriptsize\thepage \hfil \leftmark\mbox{}}%
\def\@oddfoot{ 978-0-7381-1102-5/20/\$31.00~\copyright~2020 IEEE\hfill 
\leftmark\mbox{}}%
\def\@evenfoot{\MYfooter}}

\pagestyle{headings}
\addtolength{\footskip}{0\baselineskip}
\addtolength{\textheight}{0\baselineskip}

\makeatletter
\newcommand{\linebreakand}{%
  \end{@IEEEauthorhalign}
  \hfill\mbox{}\par
  \mbox{}\hfill\begin{@IEEEauthorhalign}
}
\makeatother

\author{
    \IEEEauthorblockN{Ali Hasan Md. Linkon}
 \IEEEauthorblockA{\textit{Computer Science and Engineering} \\
 Shahjalal University of Science\\ and Technology\\
 Sylhet, Bangladesh\\
  linkon3.1416@gmail.com}
  \and
 \IEEEauthorblockN{Md. Mahir Labib}
 \IEEEauthorblockA{\textit{Computer Science and Engineering} \\
 Shahjalal University of Science\\ and Technology\\
 Sylhet, Bangladesh\\
 mdmahirlabib@gmail.com}
 \and
 \IEEEauthorblockN{Faisal Haque Bappy}
 \IEEEauthorblockA{\textit{Computer Science and Engineering} \\
Shahjalal University of Science\\ and Technology\\
Sylhet, Bangladesh\\
hbfaisal66@gmail.com}
  \linebreakand 
\IEEEauthorblockN{Soumik Sarker}
\IEEEauthorblockA{\textit{Computer Science and Engineering} \\
Shahjalal University of Science\\ and Technology\\
Sylhet, Bangladesh\\
ronodhirsoumik@gmail.com}
\and

\IEEEauthorblockN{Marium-E-Jannat}
\IEEEauthorblockA{\textit{Computer Science and Engineering} \\
Shahjalal University of Science\\ and Technology\\
Sylhet, Bangladesh\\
jannat-cse@sust.edu}
\and 

\IEEEauthorblockN{Md Saiful Islam}
\IEEEauthorblockA{\textit{Computer Science and Engineering} \\
Shahjalal University of Science\\ and Technology\\
Sylhet, Bangladesh\\
saiful-cse@sust.edu}
}

\maketitle

\begin{abstract}
Automatic detection and recognition of banknotes can be a very useful technology for people with visual difficulties and also for the banks itself by providing efficient management for handling different paper currencies. Lightweight models can easily be integrated into any handy IoT based gadgets/devices. This article presents our experiments on several state-of-the-art deep learning methods based on Lightweight Convolutional Neural Network architectures combining with transfer learning. ResNet152v2, MobileNet, and NASNetMobile were used as the base models with two different datasets containing Bangladeshi banknote images. The Bangla Currency dataset has 8000 Bangladeshi banknote images where the Bangla Money dataset consists of 1970 images. The performances of the models were measured using both the datasets and the combination of the two datasets. In order to achieve maximum efficiency, we used various augmentations, hyperparameter tuning, and optimizations techniques. We have achieved maximum test accuracy of  98.88\% on 8000 images dataset using MobileNet, 100\% on the 1970 images dataset using NASNetMobile, and 97.77\% on the combined dataset (9970 images) using MobileNet. 

\end{abstract}


\section{Introduction}
Bangladeshi banknote detection is vital from many points of view. Banknote detection technology can benefit many people with visual disabilities. According to WHO, at least 2.2 billion people worldwide are affected by vision impairment or blindness\cite{WHO}. In Bangladesh, approximately 0.8 million people are blind\cite{Guardian}. Visually impaired people may use this form of technology to make better decisions about differences in notes. ATMs or other currency detection devices can also use the banknote detection model for better improvement. However, the lack of a dataset and efficient model reduces Bangladeshi banknote detection's chance of improvement. Due to the lack of resources, limited research work was done on Bangladeshi banknote detection. In order to avoid this problem, data augmentation and transfer learning can be a possible solution. In recent times a dataset of 1970 Bangla banknote images has been available on Kaggle named `Bangla Money (Taka recognition dataset)'\cite{NoushadSojib}. Moreover, Murad et al.\cite{HasanMurad} has also created a vast dataset of 8000 banknote images. 
 
\begin{figure}
\centering
\includegraphics[width=9cm]{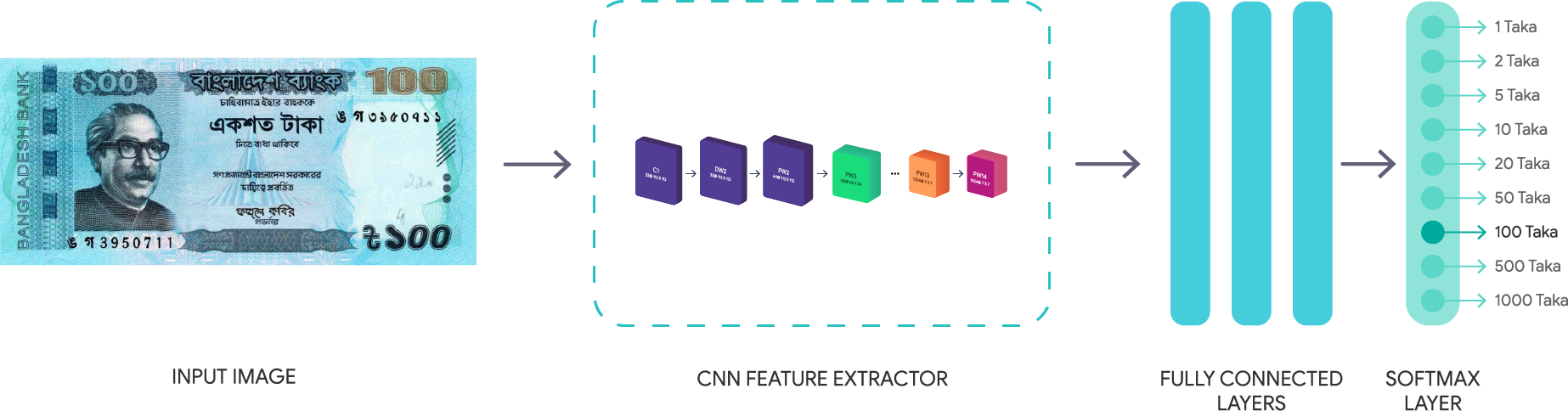}
\caption{Simplified architecture of our model}
\label{fig:Model}
\end{figure}

In this paper, we aim to investigate transfer learning combining with deep learning methods for banknote detection and recognition. Transfer learning has helped us to overcome the limitations of the dataset. We have trained two datasets using MobileNet, NASNetMobile, and ResNet152v2 pre-trained on the ImageNet dataset. We have also used various augmentation techniques, such as rotating, zooming, shifting, and shearing, to reduce the dataset's limitations. In order to achieve maximum efficiency, we have tuned hyperparameters and used different optimization techniques like reducing learning rates on plateaus, early stopping, and model checkpoints. We have achieved the best result of 98.88\% in 8000 images dataset using MobileNet, 100\% in the 1970 images dataset using NASNetMobile, and 97.77\% in the combined dataset using MobileNet. Figure \ref{fig:Model} represents the simplified architecture of our model. In the ‘CNN feature extractor’ section, we have used different models such as MobileNet, NASNetMobile, and ResNet152v2. We have shown that lightweight models like MobileNet and NasNetMobile have achieved better accuracy than the heavyweight model ResNet152v2.

Section \ref{RW} addresses some notable related works on automatic currency recognition and detection system. In Section \ref{Dataset}, we have included a concise explanation of our experimented datasets. We have discussed our proposed method in the Section \ref{Methodology}. In Section \ref{result}, we have compared and analyzed the received results and accuracy of different models. In Section \ref{Limitations}, we have discussed the limitations and scope of potential research in Bangladeshi banknote recognition.

\section{Related Work} \label{RW}
Nadim Jahangir et al.\cite{NadimJahangir} has contributed significantly to the early stage of Bangladeshi banknote recognition. According to them, they first converted images to gray-level images and passed through Histogram Equalization. The equalized image then was converted to a binary image by thresholding. Various MASKs were imposed on the binary image to get slab values and make it readable to ANN. They claimed their model's average accuracy as 98.57\% with incorrect recognition for some old, defective, taint, and worn-out notes.  

In M.M. Rahman et al.\cite{M.M.Rahman}, the dataset was collected from various sources over the Internet. They applied Oriented FAST and Rotated BRIEF (ORB) on that image data. They claimed their system could identify Bangladeshi paper currency notes with 89.4\% accuracy on the white background and 78.4\% accuracy on the complex background.

Hasan Murad et al.\cite{HasanMurad} contributed a vast novel dataset of 8000 images of Bangladeshi banknotes. For the classification of banknotes, they used MobileNet\cite{MobileNet} deep learning architecture. They claimed to achieve an overall accuracy of 99.80\% in the testing phase. However, their model recognized any background other than a white one as a valid note as their dataset did not cover all background variations.

In Shubham Mittal et al.\cite{ShubhamMittal}, they used the transfer learning technique on 380 images dataset of  Indian Currency Rupee. Pre-trained lightweight MobileNet\cite{MobileNet}  achieved 96.6\% accuracy on the test dataset.

Abhishek Pathak et al.\cite{AbhishekPathak} introduced a mobile-based Indian currency detection model that would allow visually impaired people to check a currency's value. K-means clustering was used for the classification. They claimed to use features like banknote color, ROI, and background.

N. A. J. Sufri et al.\cite{N.A.J.Sufri} used the banknote region and orientation of using Malaysian Ringgit banknotes. According to their claim, both k-NN and Decision Tree Classifier achieved 99.7\% accuracy while SVM and Bayesian Classifier achieved 100\% accuracy.

\begin{figure}
\centering
\includegraphics[scale=0.25]{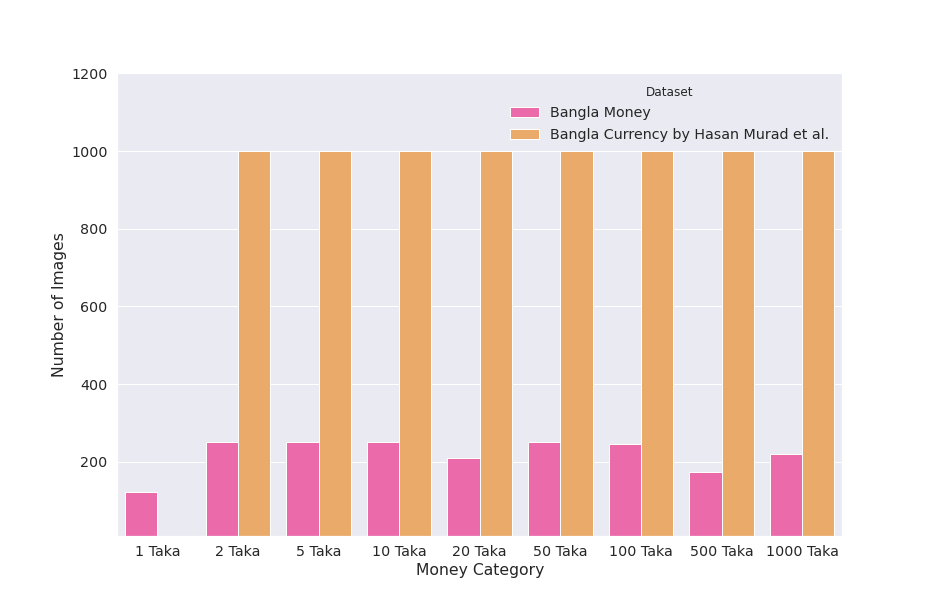}
\centering
\caption{Comparison between the Datasets, Bangla Money (Taka recognition dataset)\cite{NoushadSojib} and Bangla Currency Dataset\cite{HasanMurad}} \label{fig:DatasetGraph}
\end{figure}

\section{Dataset}\label{Dataset}

For this study, we have collected datasets from two different sources. One is Kaggle's open-access dataset named as `Bangla Money (Taka recognition dataset)'\cite{{NoushadSojib}}, and another is from Hasan Murad et al.’s[8] with the author's approval. `Bangla Money (Taka recognition dataset)'\cite{{NoushadSojib}} by Noushad Sojib et al. consists of 1970 Bangladeshi banknote images of 9 different categories. This dataset has 1, 2, 5, 10, 20, 50, 100, 500, 1000 Taka Bangla banknotes. All images are captured with mobile phone camera, and each image is of 120$\times$250 pixels.

Hasan Murad et al.'s\cite{HasanMurad} `Bangla Currency Dataset' consists of 8000 images of Bangladeshi banknotes. Currently, this is the largest Bangla banknote dataset to our best knowledge. However, this dataset has not contained any 1 Taka Bangladeshi banknote. This dataset has data of 8 classes. For each class, this dataset has 1000 images. For making the dataset adjustable to real-life scenery, all of the images are taken in arbitrary lighting, resolution, folded, and background. Each of the image sizes is 224$\times$224 pixels. Figure \ref{fig:DatasetGraph} represents the number of images and categories in the two datasets. Figure \ref{fig:DatasetImageSample} denotes sample images from both datasets.

As both of the dataset's size is not large enough, data augmentation is essential for the model improvement. Due to proclaimed efficiency, we have used online data augmentation to reduce overfitting. Each image is augmented into 10 more images. As a result, the number of images to be trained increased 10 times. Augmentations techniques are given below:
\let\labelitemi\labelitemii
\begin{itemize}
  \item Rotating (range = [0,180])
  \item Width \& Height shift (range = 0.1)
  \item Shearing (range = 0.1)
  \item Zooming (range = [0.8, 1.5])
  \item Horizontal flip
  \item Fill mode (nearest)
\end{itemize}

\section{Methodology}
\begin{figure}
\centering
\includegraphics[scale=0.35]{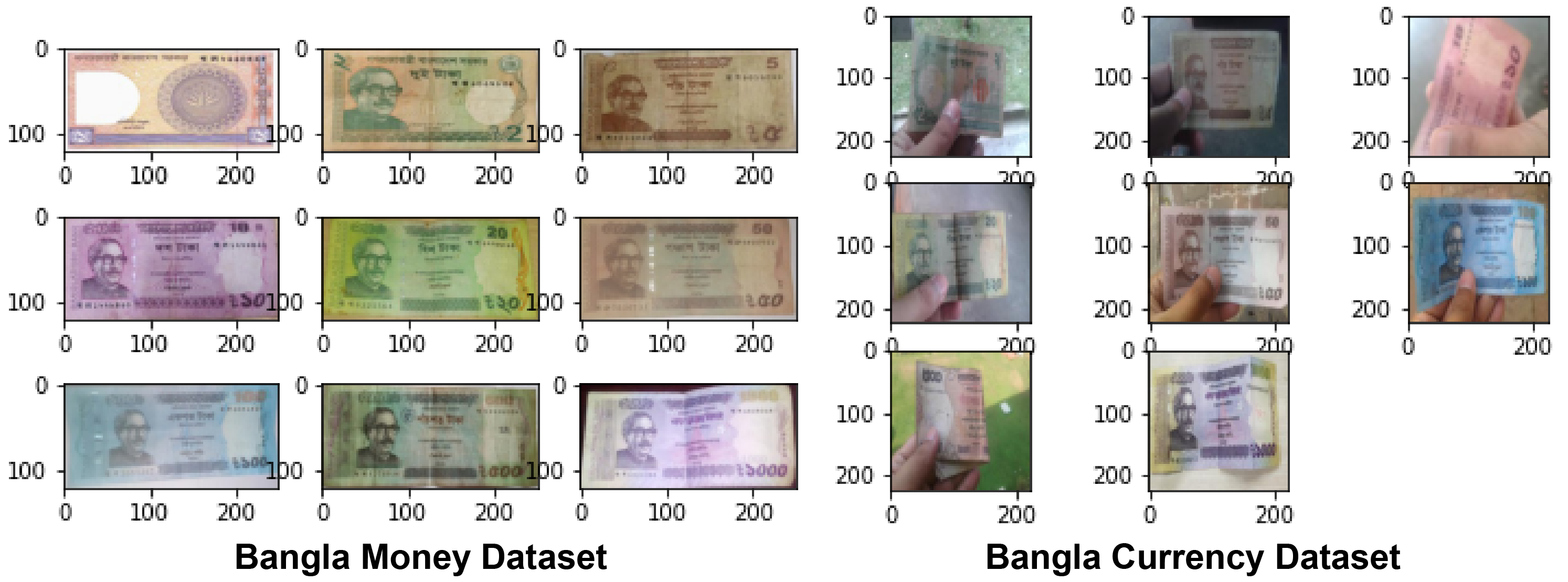}
\centering
\caption{Sample images from Bangla Money Dataset and Bangla Currency Dataset by Hasan Murad et al.}
\label{fig:DatasetImageSample}
\end{figure}

\label{Methodology}

In the Transfer Learning process, a computer can intelligently apply previously learned knowledge to solve new problems faster and efficiently. Suppose a deep learning model, trained to detect a cat can be used to detect a dog. If this model is used as a starting point and retrained to detect a dog, we can say transfer learning is applied. Fine-tuning is an essential aspect of transfer learning. The hyperparameters used for the base model with base dataset may not work for a different dataset. In fine-tuning, hyperparameters are selected in such a way so that the pre-trained model works with the new dataset. In our case, we have used 3 state-of-the-art models: MobileNet, ResNet152v2, and NASNetMobile. The models are pre-trained using `ImageNet' weights.

\subsubsection{\textit{MobileNet}\cite{MobileNet}}
In MobileNet, Point convolution and Depthwise convolutions are used. Point convolution maps an input pixel with an output pixel in all the channels used.

\subsubsection{\textit{ NASNetMobile}\cite{NASNet}}
NASNetMobile has two types of convolutional cells used after one another multiple times. They are normal and reduction cell. Both these return a feature map.

\subsubsection{\textit{Resnet152v2}\cite{ResNet}}
ResNet uses a skip-connection, which helps the gradient to backpropagate and train deeper networks. ResNet has two main types of blocks; identity block and Convolutional block.

All the models are pre-trained with various features from `ImageNet' dataset and require an input image size of $224\times224\times3$. We have set 50 epochs for training. But we get a convergence of our models at random epochs, as shown in figure \ref{fig:Accuracy}. We have used a batch size of 32 for our training. Using \textit{Adam optimizer}, we have chosen a learning rate at 0.0001. We have decreased the learning rate on plateaus and saved the best model based on lowest validation loss. As a result of which, the expected results are obtained within very few epochs. To avoid underfitting or overfitting, we have set a monitor so that after every 2 epochs, if the validation loss does not decrease, we decreased the learning rate by a factor of 0.8.

We have used 4 fully connected layers. Since ImageNet has 1000 categories for classification, we have used 1024 neurons for first dense layer. Sequentially, the next dense layers have 512, 512, 256, and 128 neurons, respectively. A \textit{softmax} classification layer with 8 classes was used for the `Bangla Currency' dataset by Hasan Murad et al.\cite{HasanMurad}, and with 9 classes was used for the `Bangla Money'\cite{NoushadSojib} dataset. \textit{Softmax} classification gives us the logits or probabilities for each category. For multi-class classification, it is wiser to choose \textit{Categorical Cross-Entropy} as probabilistic loss function. 

\section{Result and Discussion} \label{result}
In this section, we will present the details of our experimental result.  All these programs were implemented using Python 3.6 and Keras framework. The experiments were carried out on Google Colaboratory using GPU as runtime. Google Colab provides Tesla K80, P-100, and Tesla T4 randomly.

\subsection{Comparison of Model’s Performance}
In Table \ref{tab:table1}, we have provided an overview of the experiments carried out on `Bangla Currency', `Bangla Money', and the combination of the two datasets. We have used Precision, Recall, Test Accuracy, and F1 score as metrics for evaluation purposes.

In the `Bangla Money'\cite{NoushadSojib} dataset with 1970 images, NASNetMobile gives the best accuracy 100\%  in the test set. In the `Bangla Currency'\cite{HasanMurad} dataset with 8000 images, MobileNet gives the best accuracy 98.88\% in the test dataset. Both of the datasets give decent accuracy and less bias \& variance on MobileNet architecture. For this reason, we have selected MobileNet architecture for the combination of the two datasets. `Bangla Currency' by Hasan Murad et al.\cite{HasanMurad} datasets contains arbitrary lighting and background. But `Bangla Money' dataset has no arbitrary background. So, the combined dataset has worse accuracy than the individual dataset. However, data augmentation and fine-tuning improve accuracy. In the combined dataset with 9970 images, MobileNet gives a decent accuracy of 97.77\% in the test set. In all of the cases, lightweight models like MobileNet and NASNetMobile are performing similar to heavyweight model like ResNet152v2.  So, these lightweight models can be easily implemented in any smartphone or handy gadgets/devices based on IoT technology.

In figure \ref{fig:Accuracy}, training and validation accuracy is shown. In some cases, we have tried to increase the learning rate or decrease it so that our model does not underfit or overfit.

\begin{table}
\caption{Comparison of performances of different datasets on different CNN architectures}
\centering
\label{tab:table1}
\begin{tabular}{|l|l|c|c|c|c|}
\hline
\textbf{Dataset} & \textbf{\begin{tabular}[c]{@{}l@{}}Model\\ (CNN)\end{tabular}} & \textbf{Precision}  &
\textbf{\begin{tabular}[c]{@{}c@{}}Test \\ Accuracy\end{tabular}} & \textbf{F1 Score} \\ \hline
 & MobileNet & 0.99 &  0.99 & 0.99 \\
Bangla Currency & ResNet152 & 0.95 &  0.95 & 0.95 \\
 & NASNetMobile & 0.93 & 0.93 & 0.92 \\ \hline
 & MobileNet & 0.99  & 0.99 & 0.99 \\
Bangla Money  & ResNet152 & 0.98  & 0.98 & 0.98 \\
 & NASNetMobile & 1.0 & 1.0 & 1.0 \\ \hline
 
Combined Dataset & MobileNet & 0.98 & 0.98 & 0.98 \\ \hline
\end{tabular}
\end{table}

\begin{figure}[]
	\centering
	\includegraphics[scale=0.26]{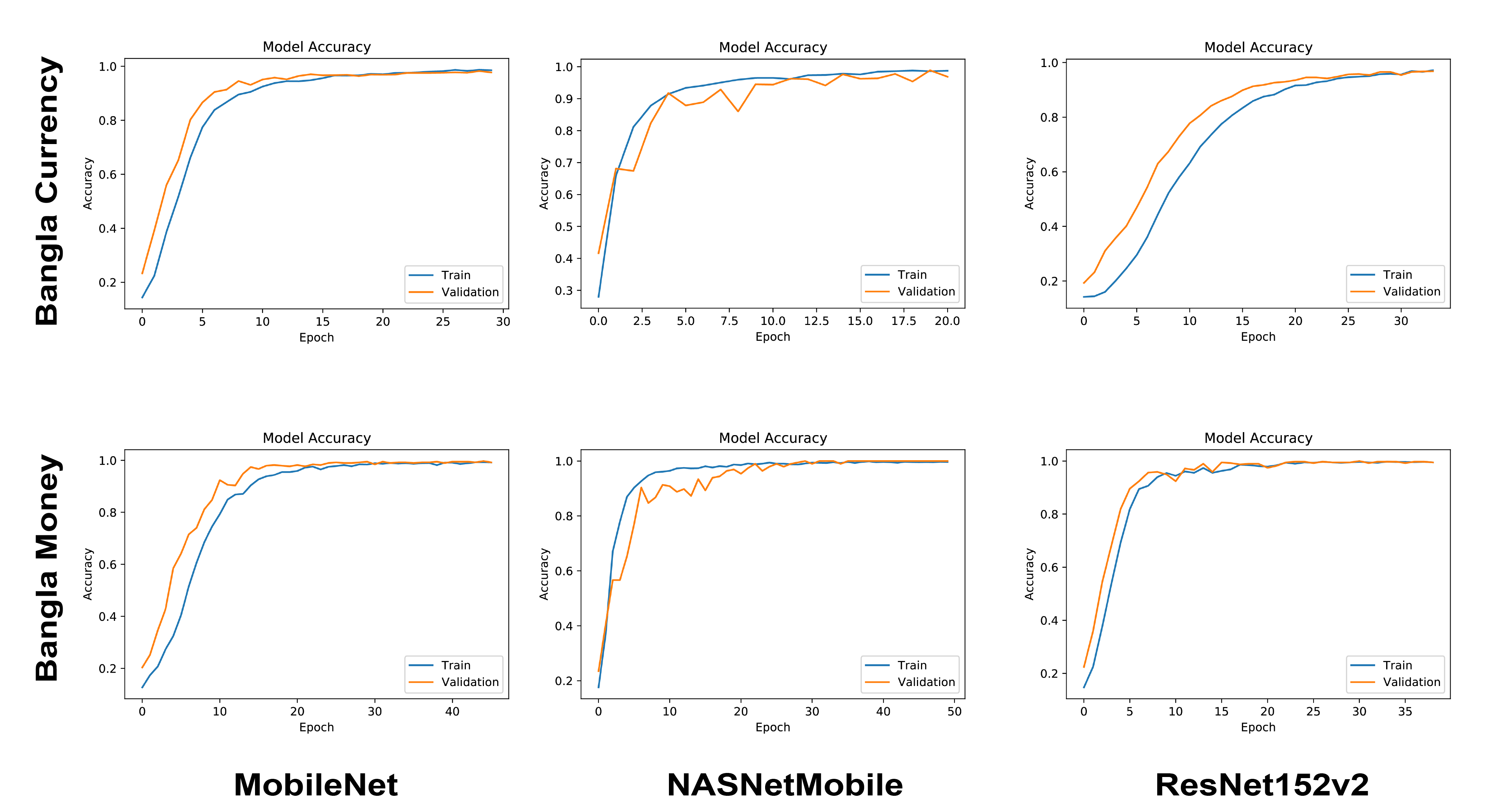}
	\caption{Comparison of training and validation accuracy of different models under two datasets}
	\label{fig:Accuracy}
\end{figure}

\subsection{Comparison with Others’ Contribution}
In table \ref{tab:compare}, we have presented a comparison of our accuracy, dataset, and approach with others who worked with Bangladeshi banknotes. There is very few application of CNN in detecting Bangladeshi banknote. We have gained  100\% accuracy on the new dataset `Bangla Money'\cite{NoushadSojib}. We have achieved 99.88\% accuracy on the `Bangla Currency' dataset introduced in the paper of Hasan Murad et al. \cite{HasanMurad}. We have also combined both of the datasets and achieved  97.77\% accuracy. With fewer epochs, we have reached almost similar scenarios as Hasan Murad et al. \cite{HasanMurad} without overfitting the model. 

\begin{table}[]
\caption{Comparison of performance and dataset with others works on Bangladeshi banknotes detection}
\centering
\label{tab:compare}
\begin{tabular}{|p{1.5cm}|p{1.5cm}|p{2cm}|p{2cm}|}
\hline
\textbf{Contributors} &  \textbf{Total images in datasets} &  \textbf{Methodology} &  \textbf{Accuracy} \\ \hline
Nadim Jahangir et al.\cite{NadimJahangir} & 1700 & ANN with Axis Symmetrical Masks & 98.57\%  \\  \hline
M.M. Rahman et al.\cite{M.M.Rahman} &  Not mentioned & Oriented FAST and Rotated BRIEF (ORB) & 78.4\% \\ \hline
Hasan Murad et al.\cite{HasanMurad}  & 8000 & Transfer learning using MobileNet & 99.80\%\\ \hline
Ours & 9970 (8000+1970) & Transfer learning using MobileNet, ResNet152v2 and NASNetMobile  &  98.88\% in 8000 images, 100\% in 1970 images and 97.77\% in 9970 images dataset\\ \hline
\end{tabular}
\end{table}

\section{Limitations and Future Directions} \label{Limitations}
The shortage of datasets is a major concern in Bangladeshi banknotes detection and recognition system. Arbitrary lighting and background are still vital issues for banknote detection. Our models struggle to perform well in several cases where images have complex background. Our main aim was to enhance model’s efficiency in the detection of Bangladeshi banknotes, which is shown in the paper. As we have selected models as lightweight, it can easily be implemented in any device such as smartphones. We have planned to integrate fake banknote detection algorithms with our lightweight model to ensure banknote security on which we are currently working. We are also working on analyzing the automatic banknote detection system’s impact on visually impaired people.  As all the handy gadgets will be based on IoT technology in the near future, so a lightweight model for banknote detection and recognition is needed. We found that our proposed model (based on the experimented models) can be a promising candidate in this regard. 

\section{Conclusion} \label{Conclusion}
In this paper, we have tried to differentiate state-of-the-art architectures with two datasets containing Bangladeshi banknote images. We have also used our best model on the combined dataset and achieved a decent result. It has shown that Transfer Learning approach plays a significant role in Bangladeshi banknotes detection. We have found out pre-trained lightweight model MobileNet, and NasNetMobile give better accuracy in detecting banknotes. These models can easily be deployed on any IoT devices for practical use.


\section*{Acknowledgment}
We would like to thank Shahjalal University of Science and Technology (SUST) and SUST NLP research group for their support.

\vspace{12pt}


\begin{thebibliography}{00}
\bibitem{WHO}
Who.int. 2020. Vision Impairment And Blindness. [online] Available at: https://www.who.int/news-room/fact-sheets/detail/blindness-and-visual-impairment [Accessed 24 July 2020].

\bibitem{Guardian}
the Guardian. 2020. Bangladesh Fights To End Blindness. [online] Available at: https://www.theguardian.com/world/2010/sep/28/bangladesh-volunteers-childhood-blindness-treatment [Accessed 24 July 2020].
\bibitem{NoushadSojib}
Sojib, N., 2020. Bangla Money. [online] Kaggle.com. Available at: https://www.kaggle.com/nsojib/bangla-money [Accessed 24 July 2020].

\bibitem{NadimJahangir}
N. Jahangir and A. R. Chowdhury, ``Bangladeshi banknote recognition by neural network with axis symmetrical masks," 2007 10th international conference on computer and information technology, Dhaka, 2007, pp. 1-5, doi: 10.1109/ICCITECHN.2007.4579423.
\bibitem{M.M.Rahman}
Rahman M.M., Poon B., Amin M.A., Yan H. (2014) Recognizing Bangladeshi Currency for Visually Impaired. In: Wang X., Pedrycz W., Chan P., He Q. (eds) Machine Learning and Cybernetics. ICMLC 2014. Communications in Computer and Information Science, vol 481. Springer, Berlin, Heidelberg
\bibitem{HasanMurad}
Hasan Murad, Nafis Irtiza Tripto, and Mohammad Eunus Ali. 2019. Developing a bangla currency recognizer for visually impaired people. In Proceedings of the Tenth International Conference on Information and Communication Technologies and Development (ICTD ’19). Association for Computing Machinery, New York, NY, USA, Article 56, 1–5. DOI:https://doi.org/10.1145/3287098.3287152

\bibitem{ShubhamMittal}
S. Mittal and S. Mittal, ``Indian Banknote Recognition using Convolutional Neural Network," 2018 3rd International Conference On Internet of Things: Smart Innovation and Usages (IoT-SIU), Bhimtal, 2018, pp. 1-6, doi: 10.1109/IoT-SIU.2018.8519888.


\bibitem{AbhishekPathak}
Pathak A., Aurelia S. (2020) Mobile-Based Indian Currency Detection Model for the Visually Impaired. In: Paiva S., Paul S. (eds) Convergence of ICT and Smart Devices for Emerging Applications. EAI/Springer Innovations in Communication and Computing. Springer, Cham

\bibitem{N.A.J.Sufri}
N. A. J. Sufri, N. A. Rahmad, N. F. Ghazali, N. Shahar and M. A. As’ari, ``Vision Based System for Banknote Recognition Using Different Machine Learning and Deep Learning Approach," 2019 IEEE 10th Control and System Graduate Research Colloquium (ICSGRC), Shah Alam, Malaysia, 2019, pp. 5-8, doi: 10.1109/ICSGRC.2019.8837068.


\bibitem{MobileNet}
Andrew G. Howard, Menglong Zhu, Bo Chen, Dmitry Kalenichenko, Weijun Wang, Tobias Weyand, Marco Andreetto and Hartwig Adam.
\newblock MobileNets: Efficient Convolutional Neural Networks for Mobile Vision Applications, 2017;
\newblock arXiv:1704.04861.
\bibitem{NASNet}  
Barret Zoph, Vijay Vasudevan, Jonathon Shlens and Quoc V. Le.
\newblock Learning Transferable Architectures for Scalable Image Recognition, 2017;
\newblock arXiv:1707.07012.
\bibitem{ResNet}
Kaiming He, Xiangyu Zhang, Shaoqing Ren and Jian Sun.
\newblock Deep Residual Learning for Image Recognition, 2015;
\newblock arXiv:1512.03385.

\end{thebibliography}
\end{document}